\documentclass[journal]{IEEEtran}

\ifCLASSINFOpdf
\else
   \usepackage[dvips]{graphicx}
\fi
\usepackage{url}
\usepackage{graphicx}
\usepackage{amsfonts}
\usepackage{amsmath}
\usepackage{caption}
\usepackage{subcaption}
\usepackage[hidelinks]{hyperref}

\begin{document}
% \maketitle
\title{DC is all you need: describing ReLU from a signal processing standpoint}

\author{Christodoulos Kechris, Jonathan Dan, Jose Miranda, and David Atienza, \IEEEmembership{Fellow, IEEE}
\thanks{This research was supported in part by the Swiss National Science Foundation Sinergia grant 193813: "PEDESITE -
Personalized Detection of Epileptic Seizure in the Internet of Things (IoT) Era", and the Wyss Center for Bio and Neuro Engineering: Lighthouse Noninvasive Neuromodulation of Subcortical Structures. }
\thanks{All authors are affiliated with the Embedded Systems Laboratory (ESL), EPFL, Switzerland.}
\thanks{Corresponding author C.K. e-mail: christodoulos.kechris@epfl.ch}}

\maketitle

\begin{abstract}
Non-linear activation functions are crucial in Convolutional Neural Networks. However, until now they have not been well described in the frequency domain. In this work, we study the spectral behavior of ReLU, a popular activation function. We use the ReLU's Taylor expansion to derive its frequency domain behavior. We demonstrate that ReLU introduces higher frequency oscillations in the signal and a constant DC component. Furthermore, we investigate the importance of this DC component, where we demonstrate that it helps the model extract meaningful features related to the input frequency content. We accompany our theoretical derivations with experiments and real-world examples. First, we numerically validate our frequency response model. Then we observe ReLU's spectral behavior on two example models and a real-world one. Finally, we experimentally investigate the role of the DC component introduced by ReLU in the CNN's representations. Our results indicate that the DC helps to converge to a weight configuration that is close to the initial random weights. 
\end{abstract}

\begin{IEEEkeywords}
Neural network, Rectified Linear Unit, Convolution. 
\end{IEEEkeywords}

\IEEEpeerreviewmaketitle

\section{Introduction}
\IEEEPARstart{C}{onvolutional} Neural Networks (CNN) consist of blocks that include linear convolutional layers, activation functions, normalization and down-sampling through pooling. The convolutional layers act as linear filters on the input signal, while the activation function introduces non-linearities to the network. Many activation functions have been proposed \cite{hendrycks2016gaussian, clevert2015fast}. The Rectified Linear Unit (ReLU) is commonly adopted due to its simple formulation and fast computation. However, its properties as a transfer function have not yet been well described \cite{watanabe2021image}. 

In \cite{watanabe2021image}, ReLU's frequency response is approximated from empirical observations on mono-frequency oscillations. In \cite{liu2022design, ayat2019spectral}, the ReLU is approximated as a quadratic function whose coefficients are selected empirically. Rahaman et al. \cite{rahaman2019spectral} use Fourier analysis to investigate neural network bias towards learning low-frequency functions. ReLU has also been studied from a probabilistic point of view by Pilipovsky et al. \cite{pilipovsky2023probabilistic}. 

A deeper understanding of the activation function can help better understand CNN's inner mechanism. Here, two key points are relevant. First, deep networks are prone to learning \textit{simpler} cues when they are informative for the given task \cite{scimeca2021shortcut}. And second, randomly initializing a network places it already \textit{close} to a locally optimal solution\cite{wu2017towards}.

In this work, we give an exact mathematical description of ReLU activations in the frequency domain. We show that ReLUs maintain the original input frequency content and additionally introduce higher frequencies and a DC component. Importantly, the latter is modulated by the frequency content of the input signal. We then show how CNNs leverage this modulation to converge to a simple solution close to the initial random weights. We accompany our theoretical findings with simulations and real-world examples. The code for reproducing our results is available here: \url{https://github.com/esl-epfl/relu_dc_is_all_you_need}.

In the remainder of the manuscript, we first formulate our problem and provide a derivation of the ReLU description in the frequency domain (Sections \ref{sec:problem_formulation} and \ref{sec:relu_in_the_frequency_domain}). We also study the DC modulation introduced by the ReLU (Section \ref{sec:dc_component_as_a_feature_extractor}). We then experimentally validate our ReLU frequency model in an example scenario and a real-world use case (Sections \ref{sec:relu_approximation_simulations} and \ref{sec:real_world_cnns}) . Finally, we empirically explore the role of the DC component in learning meaningful features (Sections \ref{sec:extracting_features_with_DC_components} and \ref{sec:a_minimum_almost_zero_training_cnn}). 

\section{Problem Formulation}
\label{sec:problem_formulation}
Let $x: \mathbb{R} \rightarrow \mathbb{R}$ a continuous time-domain signal, and $X(f) = \int x(t)e^{-i2\pi f t}dt$ its Fourier transform. Although, in practice, ReLU is applied on discrete signals, here we first consider the continuous-time case. We apply the ReLU operation, $ReLU: \mathbb{R} \rightarrow \mathbb{R^+}$, on $x(t)$, $y(t) = ReLU(x(t))$, defined as:

\begin{equation} \label{eq:relu_definition}
    y(t) = max(0, x(t)), \quad t \in \mathbb{R}
\end{equation}

We seek to characterize the Fourier transform of $y(t)$, $Y(f)$. More specifically, we describe $Y(f)$ in terms of the spectral content of $X(f)$.

\section{ReLU in the Frequency Domain}
\label{sec:relu_in_the_frequency_domain}
We can rewrite eq. \ref{eq:relu_definition} as:
\begin{equation} \label{eq:relu_sqrt_definition}
    y(t) = \frac{x(t) + |x(t)|}{2} = \frac{x(t) + \sqrt{x^2(t)}}{2}
\end{equation}
From eq. \ref{eq:relu_sqrt_definition} observe that the spectral content of $y(t)$ is $x(t)$ plus the additional terms introduced by $\sqrt{x^2}$. Without loss of generality, $x(t)$ is expressed as a sum of cosine zero-phase oscillations: $x(t) = \sum a_i cos(2 \pi f_i t)$. The following findings can be expanded for the non-zero phase. Then $x^2(t)$ is:

\begin{align}
    x^2(t) &= \left(\sum a_i cos(2 \pi f_i t)\right)^2 \nonumber \\
           &= \sum a^2_i cos^2(2\pi f_i t) \\
           &+ 2 \sum_i \sum_j a_i a_j cos(2 s\pi f_i t)cos(2 \pi f_j t) \nonumber \\
         &= sA \left( 1 + g(t) \right) = A \left( 1 + m(t) \right)
\end{align}where $A = \sum \frac{a^2_i}{2}$, $s$ is selected such that $|g(t)| < 1\ \forall t$, $g(t) = \frac{1}{s}m(t) + \frac{1-s}{s}$ and 
\begin{align}
\label{eq:mt_definition}
    m(t) &= \frac{1}{2A} \sum a^2_i cos(2\pi \cdot 2 f_i \cdot t)  \nonumber \\
         &~+ \frac{1}{A} \sum_i \sum_j a_i a_j cos(2 \pi (f_i + f_j) t)  \nonumber \\
         &~+ \frac{1}{A} \sum_i \sum_j a_i a_j cos(2 \pi (f_i - f_j) t)
\end{align} Eq. \ref{eq:relu_sqrt_definition} can be reformulated as:

\begin{align*}
    y(t) &= \frac{x(t) + \sqrt{x^2(t)}}{2} = \frac{x(t) + \sqrt{A \left( 1 + g(t) \right)}}{2}\\
         &= \frac{1}{2} x(t) + \frac{\sqrt{As}}{2} \sqrt{1 + g(t)}
\end{align*}

Although $m(t)$, eq. \ref{eq:mt_definition}, can be studied in the frequency domain, studying $\sqrt{1 + g(t)}$ is not straightforward. To expand on the terms introduced by the $\sqrt{\dots}$ we calculate its Taylor expansion around $g(t) = 0$ as:

\begin{equation} \label{eq:sqrt_sum_expansion}
    \sqrt{1 + g(t)} = \sum_{n = 0}^{\infty} c_n g^n(t)= \sum_{n = 0}^{\infty} \frac{(-1)^n (2n)!}{(1 - 2n)(n!)(4^n)}g^n(t)
\end{equation}

We can now expand $g^n(t) = \left(\frac{1}{s}h(t) + \frac{1-s}{s} \right)^n$:
\begin{equation}
    g^n(t) = \left(\frac{1}{s}\right)^n \sum_{k = 0}^{n} {k \choose n} m^k(t)\left(\frac{1-s}{s}\right)^{k-n}
\end{equation} which yields
\begin{equation}
    |x(t)| = \sqrt{sA} \sum_{n = 0}^{\infty} c_n \sum_{k = 0}^{n} \left(\frac{1}{s}\right)^k{n \choose k} m^k(t)\left(\frac{1-s}{s}\right)^{n-k}
\end{equation} and equivalently

\begin{align*}
    |x(t)| &= \sqrt{sA} \sum_{k = 0}^{\infty} m^k(t) \sum_{n = k}^{\infty} c_n {n \choose k} \left(\frac{1}{s}\right)^k \left(\frac{1-s}{s}\right)^{n-k} \\
    &= \sqrt{sA} \sum_{k = 0}^{\infty} c_k(s) m^k(t)
\end{align*}

Finally, the ReLU output, $y(t)$, can be expressed:
\begin{equation}\label{eq:relu_sqrt_sum_expansion}
    y(t) = \frac{1}{2} x(t) + \frac{\sqrt{2}}{4} \sqrt{s\sum a_i^2} \sum_{k = 0}^{\infty} c_k(s) m^k(t)
\end{equation}

Observe that $m(t)$ is composed of components at frequencies $2\cdot f_i$, and combinations of all frequencies $f_i$. Consequently, $m^n(t)$, is composed of components at frequencies multiples of $f_i$, their combinations $f_i + f_j, f_i - f_j$ and additional linear combinations of all frequency components available in $m(t)$. Additionally, raising the $cos(\cdots)$ terms of $m(t)$ in even powers introduces DC components. Hence, the ReLU operation introduces an additional DC component whose amplitude is dependent on the amplitude of all oscillations present in the input signal (see Appendix \ref{sec:appendix_dc_terms}). Finally, if $x(t)$ is constant, the output of the ReLU is also constant. 

Although the sum in eq. \ref{eq:sqrt_sum_expansion} is infinite, it converges exponentially. Hence, higher frequencies contribute minimally to $y(t)$, eq. \ref{eq:relu_sqrt_definition}. We provide a detailed proof of exponential convergence in Appendix \ref{sec:appendix_exponential_sum_convergence}. 

As a corollary, the expansion of the bandwidth of $y(t)$ caused by a single ReLU operation is limited. Although a single ReLU will introduce higher frequencies, the power of these higher frequencies is quickly reduced. Hence, in practice the resulting signal is still band-limited. Multiple consecutive ReLU layers do not add additional higher frequencies as: $y(t) = relu(relu(x(t))) = relu(x(t))$. The introduction of new frequencies throughout the network is a consequence of the combination of convolutional layers that reintroduce negative values in combination with the ReLU activation. Additionally, pooling operations band-limit the content of future activations.

To further explore these interactions between ReLU and convolutions we consider two prototypical convolutional networks. In Section \ref{sec:real_world_cnns} we link these networks with real-world ones. 

Consider the convolutional model: $h_{dif}(x) = relu(w \ast relu(w \ast (relu(...x))))$, where the convolution weights $w$ are the same for all layers and are set so that they perform discrete differentiation. Then, the differentiation output, $w \ast x$, maintains the same spectral content as the input signal $x$, while ReLU increases the input bandwidth. With enough layers $h_{dif}(x)$ will fill the entire available frequency spectrum with oscillations. 

The second example network $h_{avg}(x)$ has the same structure as $h_{dif}(x)$, but this time the weights $w$ are selected such that they perform low-pass filtering, i.e. moving average. Then, although ReLU expands the frequency range, each convolution restricts it to the bounds set by the low-pass filter. Similar behavior can be obtained by pooling, reducing the sampling frequency and, by extension, the Nyquist frequency.

\section{DC Component as a Feature Extractor}
\label{sec:dc_component_as_a_feature_extractor}

Global Average Pooling is often used after the last convolution layer to extract a feature vector. This effectively constructs a vector of DC components present in the convolution channels. We now investigate these components and demonstrate how a CNN can use them to classify signals based on their different principal frequencies. 

We define a single-layer, single-kernel network: $y(t) = ReLU(w \ast x(t))$, with $x(t) = \sum a_i cos(2 \pi f_i t)$. The output of the convolution can be expressed as $w \ast x(t) = \sum b_i a_i cos(2 \pi f_i t + \phi_i)$, with $b_i = \| \sum_{n = 0}^{M} w_n e^{-i 2 \pi f_i n}\|$ the weight of the filter at each frequency $f_i$, and similarly $\phi_i$ is the phase introduced by the filter. After passing $w \ast x(t)$ through ReLU activation, a DC component is introduced following Eq. \ref{eq:relu_sqrt_sum_expansion}:

\begin{equation}\label{eq:dc_component_equation}
    DC = \mathbb{E}[y(t)] = \frac{\sqrt{2}}{4} \sqrt{s\sum a_i^2} \cdot \sum_{k = 0}^{\infty} c_k(s) \mathbb{E}\left[m^k(t)\right]
\end{equation}

The DC component is a function of the oscillations present in the input signal $x(t)$ parameterized by the coefficients $b_i$ of the filter $w$: $DC_w(\boldsymbol{f})$, with $\boldsymbol{f}$ the vector of frequency components present in the model's input signal $x(t)$. Of note, the DC of a signal is easily extracted by calculating its average value, for example, with a global average pooling layer after a series of convolutions and ReLUs. 

As an example, take the simplest case of a single-component sinusoidal signal $x_i(t) = cos(2 \pi f_i t)$ and the task of discriminating $x_i(t)$ based on $f_i$. Then eq. \ref{eq:dc_component_equation} is be reduced to $DC(f_i) = b_i / \pi$ \footnote{For a single component input the DC component can also be calculated by calculating the integral over one period T: $\frac{1}{T}\int |y(t)|dt$. This requires knowing the region where $x(t) < 0$ which is not trivial for multi-component signals.}. Consequently, choosing any filter $w$ such that its coefficients $b_i$ are different for the different frequencies of interest $f$  is enough to classify $x_i(t)$. Parameters $b_i$ can be set by training or random initialization. We elaborate more on this in Section \ref{sec:a_minimum_almost_zero_training_cnn}.

\section{Experiments}
\subsection{ReLU Approximation Simulations}
\label{sec:relu_approximation_simulations}
As an example, let the input signal be $x(t) = \sum_{i = 1}^{4} cos(2 \pi i f_0 t)$, with $f_0 = 5Hz$. We choose a sufficiently high sampling rate ($f_s = 1024Hz$) to avoid any effects of aliasing after the ReLU operation. For the approximation calculation, we used a scaling factor $s = 20.0$ and estimated the first 100 terms for the approximation of eq. \ref{eq:relu_sqrt_sum_expansion}.

\begin{figure}[h]
    \centering
    \includegraphics[width=0.5\textwidth]{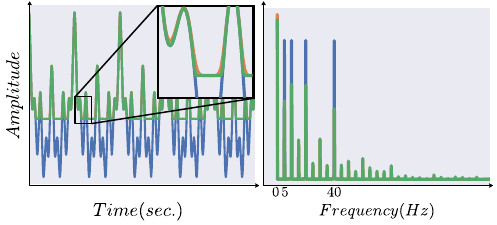}
    \caption{Time (left) and frequency (right) domain representations of the input signal (blue), ReLU (green) and ReLU approximation (orange), eq. \ref{eq:sqrt_sum_expansion}. For this signal the first 100 terms of eq. \ref{eq:relu_sqrt_sum_expansion} are sufficient for a good approximation (0.69 Relative Root Mean Squared Error).}
    \label{fig:relu_approximations}
\end{figure}

\begin{figure}[h]
    \centering
    \includegraphics[width=0.5\textwidth]{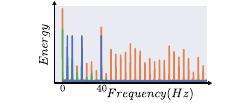}
    \caption{Frequency domain of the input signal (blue) and the outputs of the networks $h_{dif}$ (orange) and $h_{avg}$ (green). Differentiation maintains the same spectral content as its input, leading to oscillations throughout the entire frequency range due to the ReLU operations. In contrast, low-passing filters the higher frequencies introduced by the ReLU leading to a frequency-bound output signal.}
    \label{fig:relu_dif_vs_relu_avg}
\end{figure}

The input signal, the ReLU, and approximation outputs are presented in Figure \ref{fig:relu_approximations}. We also present the outputs of the convolution networks from Section \ref{sec:relu_in_the_frequency_domain}, $h_{dif}(x)$ and $h_{avg}(x)$ when they process $x(t)$. The outputs of the two networks along with the input signal are presented in Figure \ref{fig:relu_dif_vs_relu_avg}. 

\subsection{Real-World CNNs}
\label{sec:real_world_cnns}
We now investigate the activation frequency content of a CNN  \cite{kechris2024kid} trained to extract heart rate from optical heart rate sensors (photoplethysmography). The input to the network is a periodic signal comprised of two components around the heart rate frequency and its first harmonic. It can be described as $x_{heart} = \sum_{i = 1}^{2} a_i cost(2 \pi \cdot i \cdot HR \cdot t)$, where $HR$ is the heart rate and $a_1 > a_2$. 

\begin{figure}[h]
     \centering
     \begin{subfigure}[b]{0.5\textwidth}
         \centering
         \includegraphics[width=\textwidth]{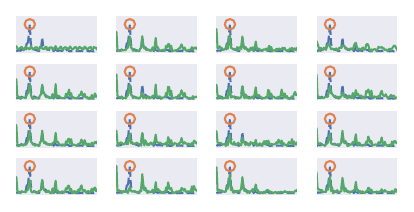}
     \end{subfigure}
     \hfill
     \begin{subfigure}[b]{0.5\textwidth}
         \centering
         \includegraphics[width=\textwidth]{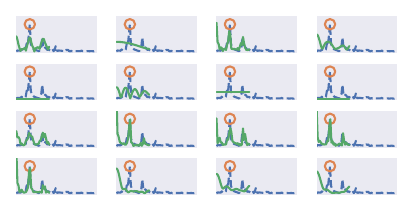}
     \end{subfigure}
        \caption{Frequency content of the activations (green) from the third (top) and sixth (bottom) convolutional layers of \cite{kechris2024kid}. The frequency components of the periodic heart signal are presented in blue, while the heart rate is indicated by an orange circle. For each layer, we plot the first 16 filters. The ReLUs introduce DC components and higher frequencies, multiples of the heart rate in the first three convolution layers. After the third, an average pooling operation reduces the available bandwidth, removing the higher frequencies. The DC component remains.}
        \label{fig:ppg_cnn_activations}
\end{figure}

The CNN is comprised of 3 convolutional blocks, each of them containing 3 ReLU convolutions followed by a pooling layer. We focus on the first two convolution blocks, studying the activations of the last layer for each block. The activations are presented in Figure \ref{fig:ppg_cnn_activations}. The first convolution/relu layers introduce the DC component and additional higher frequency components that are multiples of the heart rate frequency. This part of the CNN acts similarly to the $h_{dif}$ example, as the activation bandwidth is expanded and the convolutions do not low-pass the signal. The pooling layer (Average Pooling), after the first three convolutions, limits the bandwidth, discarding the higher frequencies introduced by the previous ReLU activations. Observe that in all activations, the DC component is prominent.

\subsection{DC Components Simplify Feature Extraction}
\label{sec:extracting_features_with_DC_components}

We examine the effect of the ReLU activation and the DC component it introduces on the training process of a feature extractor of a CNN. We do so by training three CNNs, $h_{relu}, h_{linear}, h_{linear_{DC}}$. We then evaluate their loss during training, as well as the distance between the initial random weights and the trained weights. Let $w_0$ the randomly initialized weights before training and $w_i$ the weights at epoch $i$, then we evaluate the progression of the Euclidean distance $d_i(w_0, w_i)$.

All three networks are comprised of two convolutional layers and a non-linear classifier of two fully connected layers. $h_{relu}$ uses ReLU activations in its convolution layers while $h_{linear}$ and $h_{linear_{DC}}$ have linear activations. Since $h_{linear}$ does not introduce the DC component at initialization, due to the lack of non-linear activations, we also train an additional network, $h_{linear_{DC}}$, wherein the DC component is manually added in the input.
We train the CNNs on an example dataset consisting of input-output pairs $(X, y)$:
\begin{equation}
    X_i = cos(2 \pi f_i t), \quad f_i \sim \mathcal{N}(\mu_{f}, 0.1)
\end{equation} where $\mu_{f} \in \{3.0, 5.0, 10.0\} Hz$ and
\begin{equation}
    y_i = \begin{cases}
        1 & f_i \sim \mathcal{N}(3.0, 0.1)\\
        2 & f_i \sim \mathcal{N}(5.0, 0.1)\\
        3 & f_i \sim \mathcal{N}(10.0, 0.1)
    \end{cases}
\end{equation}
For training the $f_{linear_{DC}}$ we also form the samples $X_{DC}$:
\begin{equation}
    X_i = cos(2 \pi f_i t) + DC(f_i),  DC(f_i) = \begin{cases}
        1 & f_i \sim \mathcal{N}(3.0, 0.1)\\
        2 & f_i \sim \mathcal{N}(5.0, 0.1)\\
        3 & f_i \sim \mathcal{N}(10.0, 0.1)
    \end{cases}
\end{equation}This way we partially simulate the effect of the DC, which is introduced by the ReLU in $h_{relu}$. In $h_{relu}$ the network can dynamically add additional DC components. All three networks are trained using the Sparse Categorical Cross Entropy loss and Adam optimizer ($lr = 10^{-3}, beta_1 = 0.9, beta_2 = 0.999$). The experiment is repeated 100 times. 

\begin{figure}[h]
    \centering
    \includegraphics[width=0.5\textwidth]{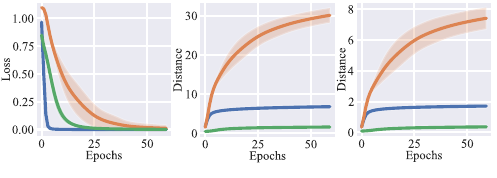}
    \caption{Training loss (left) and weight distances for the first layer (middle) and second layer (right) during training of the three CNNs: $h_{relu}$ (blue), $h_{linear}$ (orange) and $ h_{linear_{DC}}$ (green). }
    \label{fig:linear_vs_relu}
\end{figure}

The training loss, along with the weight distance, $d_i(w_0, w_i)$, for the two layers are presented in Figure \ref{fig:linear_vs_relu}. It is easier for the linear CNN to converge to a solution when the frequency-related DC component is manually added (green line in the Figure). Our analysis indicates that this is because the initialization of the weights is already close enough to a locally optimal solution. The ReLU activation provides this capability of frequency-modulated DC, hence enabling the model to rapidly converge to a solution close to the original random state. 

\subsection{A Minimum (almost) Zero-Training CNN}
\label{sec:a_minimum_almost_zero_training_cnn}

We now demonstrate how the DC can help converge to a good solution close to the initial random weights. Following the remarks in Section \ref{sec:dc_component_as_a_feature_extractor} we construct a minimal convolutional network to classify sinusoidal signals based on their principal frequency component, similar to Section \ref{sec:extracting_features_with_DC_components}. We use a single-neuron, single-filter convolution layer followed by a ReLU activation and a Global Average Pooling layer to extract the DC component. We employ this minimal CNN to classify the samples $X$, from the dataset $(X, y)$ introduced in Section \ref{sec:extracting_features_with_DC_components}.

To restrict the degrees of freedom of the convolution's kernel frequency response, we use a kernel size of two samples. The convolutional layer is randomly initialized, and no further training is used.

\begin{figure}[h]
    \centering
    \includegraphics[width=0.5\textwidth]{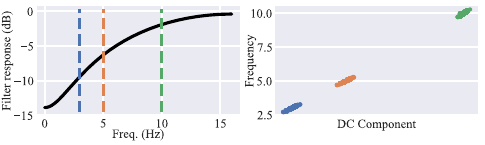}
    \caption{\textbf{Left}: Frequency response of the randomly initialized weights of the convolution. The frequencies for each class are also plotted. Each frequency corresponds to a different $b_i$, hence the initial convolution weights are good enough to classify the signals based on their frequency content. \textbf{Right:} Network output (DC) vs the input frequency for each of the three classes of signals. Each class is portrayed with a different color.}
    \label{fig:minimal_network}
\end{figure}

Figure \ref{fig:minimal_network} presents the frequency response of the convolutional layer in the frequency domain and the frequencies of the three classes.

\section{Conclusion}
In this article, we have introduced an analytical description of the ReLU activation in the Fourier domain. Our model indicates that ReLU introduces a DC component, along with high frequencies, which expands the frequency bandwidth of the input signal. In our experiments, we have shown how these theoretical remarks are found in real-world CNNs. Furthermore, we have explored the effect of the DC component introduced by ReLU on the learned features. Our results indicate that the DC helps to converge to a weight configuration that is close to the initial random weights. 

\bibliographystyle{ieeetr} % We choose the "ieee" reference style
\bibliography{refs} % Entries are in the refs.bib file

\appendices

\section{DC Terms}
\label{sec:appendix_dc_terms}
From eq. \ref{eq:dc_component_equation} the DC component is approximated as the sum: $C(A) \sum c_k(s) \mathbb{E}[m^k(t)]$, where $C(A)$ is a constant dependent only the signal amplitude. To build intuition on the signal characteristics which contribute to the DC component, we calculate the first three terms $\mathbb{E}[m^k(t)]$. We show that the DC is solely dependent on the amplitudes of the oscillations comprising the input signal $x(t)$.

\paragraph{$\boldsymbol{k = 0}$} Trivially $\mathbb{E}[m^k(t)] = 1$, contributing a constant DC component that is dependent on the amplitude of the input signal.
\paragraph{$\boldsymbol{k = 1}$} Notice that $m(t)$ is comprised only of terms $cos(\cdots)$ without any bias, eq. \ref{eq:mt_definition}, hence $\mathbb{E}[m^k(t)] = 0$.
\paragraph{$\boldsymbol{k = 2}$} We write $m(t) = \sum b_k cos(2 \pi f_k t)$, where $b_k, f_k$ the appropriate amplitudes and frequencies, e.g. for the first sum term: $b_k = a_i^2/A$ and $f_k = 2f_i$. We also consider only positive $f_k> 0$, since the sign can be absorbed by the $cos(\cdots)$. Then:
\begin{align*}
    m^2(t) &= \left(\sum_k b_k cos(2 \pi f_kt)\right) \\
           &= \sum_k \sum_l b_k b_l cos(2 \pi f_kt) cos(2 \pi f_lt)\\
           &= \frac{1}{2} \sum_k \sum_l b_k b_l \left(cos(2 \pi (f_k - f_l)t) + cos(2 \pi (f_k +f_l) t)\right)
\end{align*} From Parseval's theorem:
\begin{align*}
    \mathbb{E}[m^2(t)] &= \lim_{T\to\infty} \frac{1}{T} \int_0^T m^2(t)dt \\
    &= \frac{1}{2} \sum_k \sum_l b_k b_l (I^1_{kl} + I^2_{kl})
\end{align*} where $I^1_{kl} = \lim_{T\to\infty} \frac{1}{T} \int_0^T cos(2 \pi (f_k - f_l)t)(t)dt$ and $I^2_{kl} = \lim_{T\to\infty} \frac{1}{T} \int_0^T cos(2 \pi (f_k + f_l)t)(t)dt$. Notice that when $f_k \neq f_l$ then $I^1_{kl} = 0$, otherwise $I^1_{kl} = 1$. Similarly $I^2_{kl} = 0$ when $f_k \neq f_l$, otherwise $I^1_{kl} = 1$. Thus: 
\begin{equation}
    \mathbb{E}[m^2(t)] = \frac{1}{2} \sum_k b_k^2
\end{equation} Going back to equation eq. \ref{eq:mt_definition} each sum will contribute the following $b_k$ terms. For the first sum comprised of double frequencies ($2f_i$): $b_i = a_i^2 / (2A)$. The two next terms of frequencies $f_i - f_j$ and $f_i + f_j$ each contribute $2 a_i a_j / A$. Finally:
\begin{equation}
    \mathbb{E}[m^2(t)] = \frac{1}{8A^2} \sum a_i^4 + \frac{4}{A^2} \sum a_i^2 a_j^2
\end{equation}

\section{Exponential sum convergence}
\label{sec:appendix_exponential_sum_convergence}
We bind the upper bound of $|x(t)|$ approximated for $K$ terms. We start with the upper bound of the terms $c_k(s)$ and then $m^k(t)$.

\paragraph{Bounding $c_k(s)$} We derive upper bounds for each term of the sum separately:
For $|c_n|$:
\begin{align*}
    |c_n| &= \left|{\frac{1}{2} \choose n}\right| = \frac{{2n \choose n}}{4^n (1-2n)}
\end{align*}

\begin{equation}
    {2n \choose n} \leq \frac{4^n}{\sqrt{\pi n}}
\end{equation}

\begin{equation}
    |c_n| \leq \frac{1}{\sqrt{\pi n} (1-2n)} \leq \frac{1}{\sqrt{\pi} n^{\frac{3}{2}}}
\end{equation}
For $n \geq 1$:
\begin{equation}
    |c_n| \leq C_1 n^{-\frac{3}{2}}
\end{equation} For ${n \choose k}$:
\begin{equation}
    {n \choose k} \leq \frac{n^k}{k!}
\end{equation} Then for the sum it holds:
\begin{equation}
    |c_k(s)| \leq \sum_{n = k}^{\infty} C_1 n^{-\frac{3}{2}} \frac{n^k}{k!} \left(\frac{1}{s}\right)^k \left(\frac{1 - s}{s}\right)^{n - k}
\end{equation} We set $l = n - k$:
\begin{align*}
    |c_k(s)| &\leq \left(\frac{1}{s}\right)^{k} \sum_{l = 0}^{\infty} C_1 (l+k)^{-\frac{3}{2}} \frac{\left(l+k\right)^{k}}{k!}\left(\frac{1 - s}{s}\right)^{l} \\
    &= C_1\left(\frac{1}{s}\right)^{k} \frac{1}{k!}\sum_{l = 0}^{\infty} \left(l+k\right)^{k-\frac{3}{2}}\left(\frac{1 - s}{s}\right)^{l}
\end{align*} Since $(l + k)^{k - \frac{3}{2}} < (2k)^{k - \frac{3}{2}}$ for $l \leq k$ and grows polynomially beyond that, we can bound by an integral:
\begin{align*}
    |c_k(s)| &\leq C_1 C_2\left(\frac{1}{s}\right)^{k} \frac{1}{k!} k^{k-\frac{3}{2}}\sum_{l = 0}^{\infty} \left(\frac{1 - s}{s}\right)^{l} \\
    &= C_1 C_2\left(\frac{1}{s}\right)^{k} \frac{1}{k!} k^{k-\frac{3}{2}} \frac{1}{1-r},\ r = \left|\frac{1-s}{s}\right|
\end{align*}

\paragraph{Bounding $m^k(t)$} $x(t)$ is a sum of sinusoids, there exists $M > 0$ such that $|m(t)| < M,\ \forall t$:
\begin{equation}
    |m^k(t)| \leq M^k
\end{equation}

We can now derive the upper bound of the series:
\begin{align*}
    \left|\sum_{k = 0}^{\infty}c_k(s) m^k(t)\right| &\leq \sum_{k = 0}^{\infty}|c_k(s) m^k(t)| \\
    &\leq C_1 C_2 \frac{1}{1-r} \sum_{k = 0}^{\infty}\left(\frac{M}{s}\right)^{k} \frac{1}{k!} k^{k-\frac{3}{2}} \\
    &\leq C_1 C_2 \frac{1}{1-r} \sum_{k = 0}^{\infty}\left(\frac{M}{s}\right)^{k} k^{-\frac{3}{2}} e^k \\
    &= C_1 C_2 \frac{1}{1-r} \sum_{k = 0}^{\infty}\left(\frac{Me}{s}\right)^{k} \left(\frac{1}{k}\right)^{\frac{3}{2}}
\end{align*} which converges for $s > Me$.

We assess convergence rate by examining the tail of the series:
\begin{equation}
    \sum_{k = K+1}^{\infty} \left(\frac{eM}{s}\right)^{k} k^{-\frac{3}{2}} \leq \int_{K}^{\infty} \left(\frac{eM}{s}\right)^{x} x^{-\frac{3}{2}}dx
\end{equation} Substituting $x = t + K$:
\begin{align*}
    &\int_{0}^{\infty} \left(\frac{eM}{s}\right)^{t+K} (t+K)^{-\frac{3}{2}}dt \\
    &= \left(\frac{eM}{s}\right)^{K}\int_{0}^{\infty} \left(\frac{eM}{s}\right)^{t} (t+K)^{-\frac{3}{2}}dt \\
    &\leq \left(\frac{eM}{s}\right)^{K}K^{-\frac{3}{2}}\int_{0}^{\infty} \left(\frac{eM}{s}\right)^{t} dt \\
    &= \left(\frac{eM}{s}\right)^{K}K^{-\frac{3}{2}} \frac{1}{log\left(\frac{s}{eM}\right)}
\end{align*}
This implies exponential convergence of the ReLU approximation.

\end{document}